\documentclass[conference]{IEEEtran}
\IEEEoverridecommandlockouts

\usepackage[draft]{hyperref}
\usepackage{algorithm}
\usepackage{array}
\usepackage{stfloats}
\usepackage{url}
\usepackage{verbatim}
\usepackage{listings}
\usepackage{makecell}
\usepackage{booktabs}
\usepackage{cite}
\usepackage{amsmath,amssymb,amsfonts}
\usepackage{algorithmic}
\usepackage{graphicx}
\usepackage{textcomp}
\usepackage{xcolor}
\usepackage{booktabs}
\usepackage{multirow}

\ifCLASSOPTIONcompsoc
\usepackage[caption=false,font=normalsize,labelfont=sf,textfont=sf]{subfig}
\else
\usepackage[caption=false,font=footnotesize]{subfig}
\fi
\def\BibTeX{{\rm B\kern-.05em{\sc i\kern-.025em b}\kern-.08em
    T\kern-.1667em\lower.7ex\hbox{E}\kern-.125emX}}

\makeatletter
\newcommand{\linebreakand}{
  \end{@IEEEauthorhalign}
  \hfill\mbox{}\par
  \mbox{}\hfill\begin{@IEEEauthorhalign}
}
\makeatother

\begin{document}

\title{A Guideline-Augmented Multi-Agent Framework for Schema-as-Code Biomedical Named Entity Recognition
}

\author{
\IEEEauthorblockN{
Songtao Li\textsuperscript{1},
Shidi Zhang\textsuperscript{1},
Jianyuan Yuan\textsuperscript{2},
Fengyu Zhang\textsuperscript{3},
}

\vspace{0.5em}

\begin{tabular}{c}
\textsuperscript{1}School of Information Science and Technology, Dalian Maritime University, Dalian, China \\

\textsuperscript{2}School of Automation, Beijing Institute of Technology, Beijing, China \\

\textsuperscript{3}School of Computer Science and Engineering, Northeastern University, Shenyang, China \\

\{devin\_lee, zhangshidi\}@dlmu.edu.cn, \\
jianyuany@bit.edu.cn, zhangfengyu@mails.neu.edu.cn \\[0.3em]

\end{tabular}
}
\maketitle

\begin{abstract}
Large language models (LLMs) have shown promising potential for biomedical named entity recognition (BioNER) through instruction following and in-context learning. However, existing LLM-based BioNER methods still face two key limitations. First, retrieved demonstrations and external biomedical knowledge provide limited support for dataset-specific annotation semantics, leaving entity boundaries, type scopes, and annotation conventions ambiguous. Second, free-form generation lacks sufficient structural control, often leading to invalid formats, hallucinated mentions, duplicated entities, and boundary errors.
To address these limitations, we propose GAMA, a guideline-augmented multi-agent framework for schema-as-code BioNER. GAMA first induces candidate annotation rules from labeled training instances and verifies them against annotated data to construct reliable dataset-specific guideline memory. Guided by these verified rules, a planning component generates ranked span-type hypotheses with rationales, and a coding component converts them into schema-constrained entity objects. A verification module then checks span grounding, type validity, and structural compliance, and performs dual-loop refinement to correct invalid or low-confidence predictions. Experiments on three widely used BioNER datasets using two different LLMs demonstrate the effectiveness of GAMA, achieving the best performance in four out of six LLM-dataset combinations. Ablation studies further validate the contributions of its key components. Code is available at \url{https://github.com/lst11111/GAMA}.
\end{abstract}


\section{Introduction}

Biomedical named entity recognition (BioNER) aims to identify domain-specific entity mentions, such as diseases, chemicals, genes, proteins, and anatomical entities~\cite{song2021deep}. As a core task in biomedical information extraction, BioNER supports many downstream applications, including biomedical relation extraction~\cite{zhang2024rethinking} and biomedical event extraction~\cite{ramponi2020biomedical}. Supervised neural models and domain-specific pretrained language models such as BioBERT~\cite{lee2020biobert} and PubMedBERT~\cite{gu2021domain} have achieved substantial progress. Nevertheless, these approaches require large-scale manually annotated corpora, which are costly due to the need for both linguistic expertise and domain knowledge.

Large language models (LLMs) offer a promising alternative, as they can perform downstream tasks using instructions and a few in-context examples without additional training~\cite{brown2020language,xu2025echogpt,zhu2026ultrag}. Recent studies have mainly explored LLM-based BioNER from two perspectives: (i) retrieving similar annotated examples as in-context demonstrations to provide instance-level evidence~\cite{ge2025retrieval,liu2025eicl}, and (ii) incorporating external biomedical knowledge, such as entity semantics or UMLS information, to improve domain understanding~\cite{bian2023inspire,zhou2024biomedical}. Despite their effectiveness, these approaches still face two major limitations. First, they provide limited support for dataset-specific annotation semantics. 
Retrieved examples usually reflect local annotation decisions, while external knowledge mainly provides general biomedical information. 
As a result, dataset-specific entity boundaries, type scopes, and annotation conventions remain ambiguous, especially when entity definitions and annotation standards vary across datasets. Second, structural control remains insufficient. Since BioNER requires each predicted entity to be grounded in the input sentence, assigned to a predefined type, and returned in a valid format, free-form LLM generation may produce invalid outputs, duplicated entities, hallucinated mentions~\cite{liu2026dualpathwaycircuitsobjecthallucination}, or boundary errors, which ordinary prompting cannot fully prevent.

To address these limitations, we present GAMA, a guideline-augmented multi-agent framework for schema-as-code BioNER. Rather than prompting a single LLM to directly produce entity predictions, GAMA decomposes the extraction process into five specialized components: Annotation Rule Induction, Rule Verification, Planning, Coding, and Verification with Dual-Loop Refinement, each responsible for a distinct stage of the task. GAMA is built upon three core principles: (1) \textbf{Explicit Guideline Modeling}, where the Annotation Rule Induction Agent summarizes dataset-specific annotation rules from training instances, and the Rule Verification Agent filters unreliable or overly instance-specific rules to construct explicit and reusable guideline memory~\cite{wang2026memguardpersistingverifiersignals,liu2026conmemcontributionawarememorylonghorizon}; (2) \textbf{Multi-Agent Decomposition}, in which the Planning Agent generates mention-type hypotheses with explanatory rationales, ranks them according to their consistency with the input sentence, the predefined label schema, and the verified guideline rules, and the Coding Agent converts the most reliable hypotheses into schema-constrained code-style entity objects; (3) \textbf{Schema-as-Code Verification}, where each predicted entity object is validated for span grounding, type correctness, and structural validity, while a dual-loop refinement procedure corrects local object-level errors or revisits lower-confidence span--type hypotheses when necessary.

Our main contributions are summarized as follows:
\begin{itemize}
  \item We introduce a guideline modeling mechanism that induces and verifies dataset-specific annotation rules, providing explicit dataset-level guidance for entity boundaries, type scopes, and annotation conventions.

  \item We propose GAMA, a guideline-augmented multi-agent framework that decomposes BioNER into rule induction, rule verification, guideline-guided planning, schema-constrained coding, and prediction verification.

  \item We develop a schema-as-code verification mechanism with dual-loop refinement, which checks span grounding, type validity, and structural compliance to reduce invalid, hallucinated, and inconsistent predictions.

\end{itemize}

\section{Related Work}
\subsection{BioNER with In-Context Learning}

BioNER remains challenging due to domain-specific terminology, complex entity boundaries, and annotation ambiguity. Traditional supervised approaches often require large-scale annotated corpora, which are expensive and time-consuming to construct in the biomedical domain. To reduce such annotation dependence, recent studies have explored in-context learning (ICL) with LLMs, where models perform entity extraction by conditioning on natural language instructions and a few contextual examples without updating model parameters. Early work focused on improving the content of prompts to enhance LLMs' understanding of the task. For instance, Hu et al.~\cite{hu2024improving} designed a clinical NER prompt framework by incorporating annotation guidelines, error-analysis-based instructions, and few-shot examples, while Ge et al.~\cite{ge2025retrieval} further investigated structured static prompts that integrate dataset descriptions, high-frequency entities, UMLS knowledge, and error feedback, demonstrating that careful prompt construction can substantially improve few-shot BioNER performance.
Beyond prompt content, other studies aimed to strengthen LLMs' domain awareness and reasoning capabilities. Bian et al.~\cite{bian2023inspire} reformulated BioNER as a two-step process of span extraction followed by entity type prediction, augmented with UMLS-based external knowledge to support accurate type determination. Similarly, Li et al.~\cite{li2024rt} introduced the RT framework, which leverages retrieved demonstrations combined with Chain-of-Thought reasoning, guiding LLMs to gradually recognize entities with higher accuracy. These works highlight that effective BioNER often requires domain-specific knowledge and reasoning beyond simple text generation. More recently, retrieval-augmented demonstration selection has been explored to improve example selection in ICL. EICL~\cite{liu2025eicl} introduces an entity-aware retrieval strategy that combines semantic similarity with domain-specific terminology overlap to select more relevant in-context examples for BioNER.

\subsection{Multi-Agent Systems in Information Extraction}

Recent advances in large language models have promoted the use of multi-agent collaboration for information extraction, where different agents are assigned specialized roles to improve structured prediction through cooperation, refinement, and verification. Wang et al.~\cite{wang2025cooperative} proposed CMAS, which employs a self-annotator, a type-related feature extractor, a demonstration discriminator, and an overall predictor to construct and filter self-annotated demonstrations for zero-shot named entity recognition. Lu et al.~\cite{lu2025crossagentie} introduced CROSSAGENTIE, a cross-type and cross-task multi-agent framework for zero-shot information extraction, where multiple agents collaborate to improve structured prediction across named entity recognition and relation extraction.  Zhang et al.~\cite{zhang2026learning} developed a generation-evaluation multi-agent framework, where a generation agent synthesizes event data and an evaluation agent assesses semantic consistency to provide reward signals for iterative optimization. These studies indicate that role specialization, agent collaboration, and verification-oriented refinement are effective for improving the reliability of LLM-based structured extraction.

\begin{figure*}[t]
	\centering
	\includegraphics[width=1\textwidth]{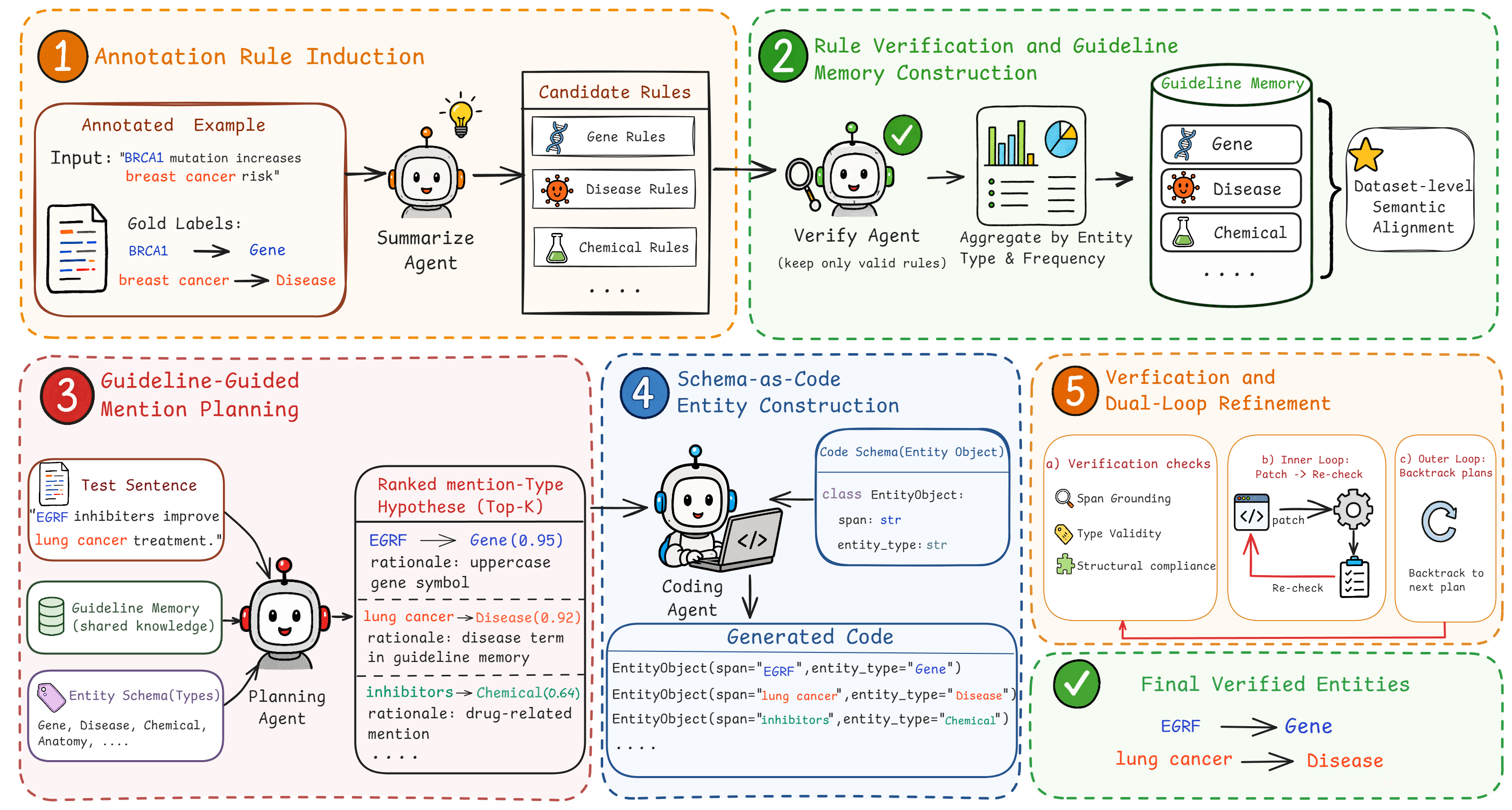}
	
	\caption{Overview of the proposed GAMA framework. The framework instantiates a guideline-augmented multi-agent paradigm for schema-as-code biomedical named entity recognition. It consists of five stages: rule induction from training annotations, guideline memory construction, guideline-guided mention planning, schema-as-code entity generation, and dual-loop verification refinement for producing final structured entities.}
	\label{fig:model}
\end{figure*}

\section{Methods}

We propose GAMA, a guideline-augmented multi-agent framework for schema-as-code BioNER. 
GAMA formulates entity extraction as a schema-constrained programming process with explicit guideline modeling and iterative verification. 
As shown in Figure~\ref{fig:model}, GAMA comprises five components: 
(1) an Annotation Rule Induction Agent for dataset-specific rule induction, 
(2) a Rule Verification Agent for rule validation and filtering, 
(3) a Planning Agent for generating ranked mention-type hypotheses, 
(4) a Coding Agent for converting hypotheses into executable entity objects, and 
(5) a Verification and Dual-Loop Refinement module for prediction validation and iterative correction.

\subsection{Annotation Rule Induction}

BioNER is strongly affected by dataset-specific annotation criteria. Different datasets may exhibit variations in entity boundary conventions, type granularity, and annotation scope, even for semantically similar biomedical mentions. Directly using input–label pairs provides only implicit supervision and fails to explicitly expose annotation semantics.

To address this limitation, we introduce an Annotation Rule Induction Agent, which transforms instance-level annotations into explicit candidate annotation rules.
Let $D^{(s)}=\{(x_i,y_i)\}_{i=1}^{M^{(s)}}$ denote the annotated training set of dataset $s$, where $x_i$ is a sentence and $y_i=\{(e_j,t_j)\}_{j=1}^{N_i}$ is its corresponding entity annotation set. Each $(e_j,t_j)$ consists of an entity mention $e_j$ and its type $t_j \in \mathcal{T}^{(s)}$.

For each training instance $(x_i,y_i)$, the rule induction agent generates a set of candidate rules:
\begin{equation}
\mathcal{R}_i^{(s)}
=
\mathcal{M}_{\mathrm{rule}}\big(P_{\mathrm{rule}}(x_i,y_i)\big),
\end{equation}
where $\mathcal{M}_{\mathrm{rule}}$ denotes the LLM-based rule generator and $P_{\mathrm{rule}}$ is the prompting function.

The global candidate rule pool is defined as:
\begin{equation}
\mathcal{R}^{(s)} = \bigcup_{i=1}^{M^{(s)}} \mathcal{R}_i^{(s)}.
\end{equation}

These rules are subsequently passed to the Rule Verification Agent for validation and filtering.

\subsection{Rule Verification and Guideline Memory Construction}

Not all induced rules are reliable or consistent with dataset-specific annotation standards. Therefore, we introduce a Rule Verification Agent to filter candidate rules before constructing the guideline memory. The key intuition is that a useful annotation rule should help recover gold entities from the training data when it is applied as an explicit constraint.

For each candidate rule $r_k \in \mathcal{R}^{(s)}$, the verification agent applies the rule to training instances and generates rule-guided predictions:
\begin{equation}
\hat{y}_{i,k}^{(s)} = \mathcal{M}_{\mathrm{ver}}\big(P_{\mathrm{ver}}(x_i, r_k)\big),
\end{equation}
where $\mathcal{M}_{\mathrm{ver}}$ denotes the verification LLM and $P_{\mathrm{ver}}$ is the verification prompt. The prediction $\hat{y}_{i,k}^{(s)}$ is then compared with the gold annotation set $y_i$. A rule is regarded as supported by an instance if it helps recover at least one gold entity with the correct mention boundary and entity type.

For each rule, we compute a support score by counting how many training instances support it. Rules with no support are discarded. The remaining rules are grouped by their associated entity types and ranked according to their support scores. For each entity type $t$, we retain the Top-$N$ verified rules to construct the dataset-specific guideline memory:
\begin{equation}
\mathcal{G}_t^{(s)} = \operatorname{Top-N}(\mathcal{R}_{t}^{\mathrm{ver},(s)}), \quad
\mathcal{G}^{(s)} = \{\mathcal{G}_t^{(s)}\}_{t \in \mathcal{T}^{(s)}} ,
\end{equation}
where $\mathcal{R}_{t}^{\mathrm{ver},(s)}$ denotes the verified rules associated with entity type $t$.

The resulting guideline memory $\mathcal{G}^{(s)}$ provides explicit and structured annotation constraints for downstream Planning, Coding, and Verification modules. By filtering rules through training-data consistency, the Rule Verification Agent reduces noisy or overly instance-specific rules without introducing additional trainable parameters.

\subsection{Guideline-Guided Mention Planning}

Given an input biomedical sentence $x$, the verified guideline memory $\mathcal{G}^{(s)}$ constructed by the Rule Verification Agent, and the entity schema $\mathcal{S}^{(s)}$, the Planning Agent generates a ranked set of mention-type hypotheses:
\begin{equation}
\mathcal{P}=\mathcal{M}_{\mathrm{plan}}\big(P_{\mathrm{plan}}(x,\mathcal{G}^{(s)},\mathcal{S}^{(s)})\big),
\end{equation}
where $\mathcal{M}_{\mathrm{plan}}$ denotes the LLM used for planning, $P_{\mathrm{plan}}$ is the planning prompt, and $\mathcal{P}$ is the generated planning set. The schema $\mathcal{S}^{(s)}$ specifies the valid entity type set $\mathcal{T}^{(s)}$ and the output constraints for dataset $s$.

The planning set is defined as:
\begin{equation}
\mathcal{P}=\{(m_i,t_i,\beta_i,\rho_i)\}_{i=1}^{K},
\end{equation}
where $m_i$ denotes a candidate biomedical entity mention, $t_i \in \mathcal{T}^{(s)}$ denotes its predicted entity type, $\beta_i \in [0,1]$ is a confidence score used for ranking, and $\rho_i$ is a natural-language rationale grounded in the dataset-specific guideline memory $\mathcal{G}^{(s)}$.

The score $\beta_i$ is used to prioritize candidate mention-type hypotheses during subsequent Coding and Verification, especially when overlapping spans or type conflicts occur. The rationale $\rho_i$ links the mention $m_i$ to relevant annotation rules in $\mathcal{G}^{(s)}$, making the type assignment explicit and providing interpretable evidence for the downstream agents.

\subsection{Schema-as-Code Entity Construction}

Existing BioNER methods often produce outputs in free-form text, tagging sequences, or loosely structured JSON objects. Such representations may lead to format inconsistencies, missing fields, or invalid entity-type assignments, making them difficult to integrate into a multi-stage verification workflow. To address this issue, we introduce the Coding Agent, which converts the ranked mention-type hypotheses from the Planning Agent into schema-constrained entity objects.

Each candidate entity is represented using a minimal BioNER schema, as illustrated in Figure~\ref{fig:entity_code}:
\begin{figure}[htbp]
    \centering
    \includegraphics[width=0.8\linewidth]{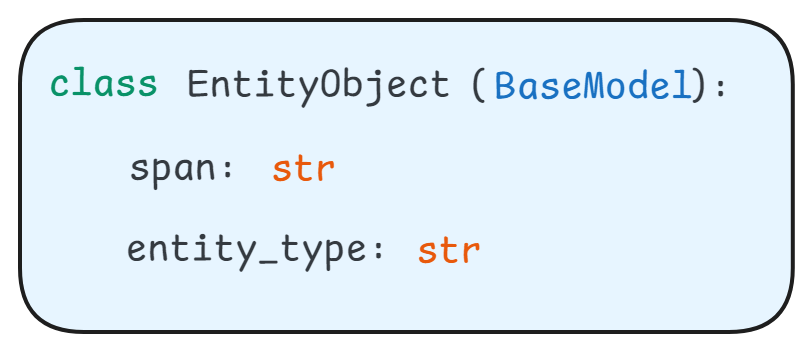}
    \caption{Code-based template for BioNER entities. \texttt{span} denotes the extracted biomedical mention, and \texttt{entity\_type} denotes its predicted category.}
    \label{fig:entity_code}
\end{figure}

Given the planning set $\mathcal{P}=\{(m_i,t_i,\beta_i,\rho_i)\}_{i=1}^{K}$, the Coding Agent instantiates each mention-type hypothesis as a schema-constrained entity object:
\begin{equation}
c_i = \operatorname{EntityObject}(m_i,t_i), \quad i=1,\ldots,K.
\end{equation}
The complete set of generated entity objects is denoted as $\mathcal{C}=\{c_i\}_{i=1}^{K}$. Here, $c_i$ corresponds to the $i$-th planning hypothesis, where $m_i$ fills the \texttt{span} field and $t_i$ fills the \texttt{entity\_type} field under the dataset-specific schema $\mathcal{S}^{(s)}$. The field \texttt{entity\_type} is constrained by the dataset-specific type set $\mathcal{T}^{(s)}$.

The ranking score $\beta_i$ and rationale $\rho_i$ are preserved as auxiliary planning information for later verification and conflict handling. Thus, the Coding Agent standardizes each mention-type hypothesis into a schema-aware entity object while maintaining the planning context needed by the subsequent Verification and Dual-Loop Refinement module.

\subsection{Verification and Dual-Loop Refinement}

After the Coding Agent generates schema-constrained entity objects, GAMA applies a prediction-level verification and refinement mechanism to ensure that the outputs are both text-grounded and schema-compliant. Each generated object is checked before being included in the final prediction set.

Given the input sentence $x$, the dataset-specific schema $\mathcal{S}^{(s)}$, and the generated entity objects $\mathcal{C}=\{c_i\}_{i=1}^{K}$, each object $c_i$ is verified along three dimensions: span grounding, type validity, and structural compliance. Span grounding checks whether the predicted mention appears in $x$, type validity checks whether the predicted type belongs to $\mathcal{T}^{(s)}$, and structural compliance checks whether the object follows the schema $\mathcal{S}^{(s)}$. Formally, this prediction-level verification is defined as:
\begin{equation}
(V_i,\epsilon_i)=\mathcal{V}_{\mathrm{pred}}(c_i,x,\mathcal{S}^{(s)}),
\end{equation}
where $V_i$ indicates whether $c_i$ passes the verification checks, and $\epsilon_i$ denotes diagnostic feedback used for refinement.

To improve robustness, GAMA employs a dual-loop refinement mechanism. The outer loop follows the ranked hypotheses produced by the Planning Agent and prioritizes candidates with higher $\beta_i$. The inner loop uses the diagnostic feedback $\epsilon_i$ to revise and re-check the corresponding entity object until it passes verification or reaches the maximum number of refinement attempts. Only verified objects are included in the final prediction set:
\begin{equation}
\mathcal{Y}^{(s)}(x)
=
\{(m_i,t_i)\mid V_i=1,\ c_i=\operatorname{EntityObject}(m_i,t_i)\}.
\end{equation}

This mechanism enables GAMA to correct object-level errors while preserving the hypothesis ranking produced by the Planning Agent, thereby producing text-grounded and schema-compliant entity predictions.

\begin{table}[t]
\centering
\caption{Statistics of the BioNER datasets used in our experiments.}
\label{tab:dataset_statistics}

\small
\renewcommand{\arraystretch}{1.10}
\setlength{\tabcolsep}{4pt}

\begin{tabular}{l c c c c}
\toprule
\textbf{Dataset} & \textbf{Labels} & \textbf{Train} & \textbf{Test} & \textbf{Entities} \\
\midrule
NCBI    & 1          & 5432  & 942  & 6881  \\
BC2GM   & 1             & 12632 & 5065 & 24583 \\
GENIA  & 5 & 16692  & 1854 & 50509 \\
\bottomrule
\end{tabular}
\end{table}

\section{Experiments}

\subsection{Datasets}

We use three public BioNER datasets in our experiments: NCBI Disease~\cite{dougan2014ncbi}, BC2GM~\cite{smith2008overview}, and GENIA~\cite{kim2003genia}. NCBI Disease focuses on disease entity recognition, BC2GM mainly contains gene and protein mentions, and GENIA includes multiple biomedical entity types. Together, these datasets cover diverse biomedical entity categories and annotation conventions, providing a comprehensive evaluation of BioNER methods. Table~\ref{tab:dataset_statistics} lists the detailed statistics.

\subsection{Baselines}

We compare our method with three representative LLM-based NER baselines, all of which perform inference without fine-tuning LLMs on task-specific data.

\begin{itemize}
    \item \textbf{GPT-NER} \cite{wang2025gpt} transforms NER into a text generation task by asking LLMs to mark entity spans with special tokens, followed by self-verification to mitigate hallucinated predictions.

    \item \textbf{CodeIE} \cite{li2023CodeIE} reformulates information extraction as a code generation task, using code-oriented LLMs to generate structured predictions in Python-style lists and dictionaries under few-shot settings.

    \item \textbf{CMAS} \cite{wang2025cooperative} performs zero-shot NER with a cooperative multi-agent framework that integrates self-annotation, type-related feature extraction, demonstration discrimination, and final prediction.
\end{itemize}







\begin{table*}[t]
\centering
\caption{Performance comparison of different methods across three BioNER datasets under different LLMs. The best results for each LLM-dataset combination are highlighted in \textbf{bold}, and the second-best results are \underline{underlined}.}
\label{tab:main_results}

\renewcommand{\arraystretch}{1.15}
\setlength{\tabcolsep}{6pt}

\begin{tabular*}{\textwidth}{@{\extracolsep{\fill}}c l c c c@{}}
\toprule
\textbf{LLM} & \textbf{Method} & \textbf{NCBI} & \textbf{BC2GM} & \textbf{GENIA} \\
\midrule

\multirow[c]{4}{*}{\textbf{DeepSeek-V4-Pro}}
& CodeIE  & 62.32 & 51.93 & \textbf{67.98} \\
& GPT-NER & 68.97 & \underline{56.18} & 64.91 \\
& CMAS    & \underline{70.54} & 52.59 & 55.50 \\
& GAMA    & \textbf{77.69} & \textbf{62.30} & \underline{65.86} \\
\midrule

\multirow[c]{4}{*}{\textbf{Gemini-3.1-Pro-Preview}}
& CodeIE  & \underline{82.30} & \underline{67.90} & \underline{69.90} \\
& GPT-NER & 81.73 & \textbf{69.97} & 68.46 \\
& CMAS    & 82.24 & 65.36 & 65.32 \\
& GAMA    & \textbf{82.34} & 67.07 & \textbf{70.06} \\

\bottomrule
\end{tabular*}
\end{table*}

\subsection{Implementation Details}

We employ DeepSeek-V4-Pro and Gemini-3.1-Pro-Preview without task-specific fine-tuning. For fair comparison, all methods use the same annotated training split: baselines select in-context examples when applicable, while GAMA uses it to induce and verify dataset-specific annotation rules. All agents collaborate through natural-language prompts and structured outputs, using the same LLM within each experiment. For each entity type, the top-$n$ verified rules are retained, where $n=10$. The Planning Agent retains only candidate entities with confidence scores $\beta_i \geq 0.9$ for subsequent Coding and Verification. The maximum number of inner-loop refinement iterations is set to $t=3$.



\section{EXPERIMENTAL RESULTS}
\subsection{Main Results}
Table~\ref{tab:main_results} presents the performance comparison of different methods across three BioNER datasets using DeepSeek-V4-Pro and Gemini-3.1-Pro-Preview. GAMA achieves the highest Micro-F1 in four out of six LLM-dataset combinations, demonstrating its effectiveness across different LLMs. These results highlight the benefits of explicit guideline modeling and schema-constrained verification for biomedical entity extraction.

Specifically, under DeepSeek-V4-Pro, GAMA outperforms the strongest baselines by 7.15 and 6.12 Micro-F1 points on NCBI and BC2GM, respectively. Under Gemini-3.1-Pro-Preview, GAMA achieves the best performance on NCBI and GENIA, surpassing the strongest baselines by 0.04 and 0.16 points, respectively. Although GAMA does not achieve the highest scores on the remaining two settings, it maintains competitive performance across different datasets. Overall, these results demonstrate the effectiveness of combining guideline memory, schema-as-code entity construction, and iterative verification for BioNER.

\subsection{Ablation Study}

Table~\ref{tab:ablation_study} evaluates five key design choices of GAMA across two LLMs and three BioNER datasets. All ablated variants consistently underperform the full model, demonstrating the effectiveness of the proposed components.

Removing Guideline Modeling leads to F1 reductions of 1.52--3.22 points, with the largest average degradation of 2.47 points across all settings, highlighting the importance of dataset-specific annotation guidelines. Removing Rule Verification causes drops of 0.87--2.65 points, demonstrating its role in improving guideline reliability. Removing Planning Rationales reduces F1 by 1.71--2.52 points, confirming their contribution to guideline-aware entity prediction.

Removing Schema-as-Code decreases F1 by 0.21--2.01 points, indicating the benefits of structured entity representations. Disabling the Verification Loop results in drops of 1.11--3.10 points, highlighting the importance of iterative prediction refinement. Overall, these findings demonstrate the complementary contributions of guideline modeling, rule verification, rationale-guided planning, structured entity representation, and iterative refinement.

\begin{table}[t]
\centering
\caption{Ablation studies on two different LLMs, evaluated on three BioNER datasets.}
\label{tab:ablation_study}
\small
\setlength{\tabcolsep}{5pt}
\renewcommand{\arraystretch}{1.10}

\begin{tabular}{lccc}
\toprule
\multicolumn{4}{c}{\textbf{DeepSeek-V4-Pro}} \\
\midrule
\textbf{Ablation Setting}
& \textbf{NCBI}
& \textbf{BC2GM}
& \textbf{GENIA} \\
\midrule
GAMA (full model)
& \textbf{77.69}
& \textbf{62.30}
& \textbf{65.86} \\

w/o Guideline Modeling
& 75.42
& 60.28
& 63.28 \\

w/o Rule Verification
& 75.12
& 61.12
& 63.21 \\

w/o Planning Rationales
& 75.17
& 60.18
& 63.52 \\

w/o Schema-as-Code
& 76.29
& 62.09
& 63.85 \\

w/o Verification Loop
& 75.96
& 60.82
& 64.12 \\
\midrule
\multicolumn{4}{c}{\textbf{Gemini-3.1-Pro-Preview}} \\
\midrule
\textbf{Ablation Setting}
& \textbf{NCBI}
& \textbf{BC2GM}
& \textbf{GENIA} \\
\midrule
GAMA (full model)
& \textbf{82.34}
& \textbf{67.07}
& \textbf{70.06} \\

w/o Guideline Modeling
& 79.12
& 63.89
& 68.54 \\

w/o Rule Verification
& 80.71
& 66.06
& 69.19 \\

w/o Planning Rationales
& 80.39
& 65.36
& 68.12 \\

w/o Schema-as-Code
& 81.10
& 66.45
& 68.86 \\

w/o Verification Loop
& 79.24
& 65.96
& 68.17 \\
\bottomrule
\end{tabular}
\end{table}

\section{Conclusions}

In this paper, we proposed GAMA, a guideline-augmented multi-agent framework for schema-as-code BioNER, addressing the limitations of insufficient annotation guidance and unconstrained entity generation. By integrating guideline induction and verification, planning and coding, and dual-loop refinement, GAMA aims to improve the accuracy and structural consistency of biomedical entity extraction. Experiments on three public BioNER datasets using two different LLMs demonstrate its competitive performance, achieving the best results in four out of six LLM-dataset combinations. Furthermore, ablation studies validate the effectiveness of its key components. These findings highlight the value of combining dataset-specific guidelines with structural constraints for reliable LLM-based BioNER. Future work will explore more efficient guideline selection and extend GAMA to broader biomedical information extraction tasks.

\bibliographystyle{IEEEtran}
\bibliography{IEEEabrv,citations}

\end{document}